\documentclass{article}
\usepackage[preprint]{spconf,amsmath,graphicx}
\copyrightnotice{
\begin{footnotesize}
\begin{tabular}{l}
     20XX IEEE. Personal use of this material is permitted. Permission from IEEE must be obtained for all other uses\\
     in any current or future media, including reprinting/republishing this material for advertising or promotional purposes \\
     creating new collective works, for resale or redistribution to servers or lists, or reuse of any copyrighted component \\ of this work in other works
\end{tabular}
\end{footnotesize}
}

\usepackage[utf8]{inputenc} % allow utf-8 input
\usepackage[T1]{fontenc}    % use 8-bit T1 fonts
\usepackage{hyperref}       % hyperlinks
\usepackage{url}            % simple URL typesetting
\usepackage{booktabs}       % professional-quality tables
\usepackage{amsfonts}       % blackboard math symbols
\usepackage{nicefrac}       % compact symbols for 1/2, etc.
\usepackage{microtype}      % microtypography
\usepackage{comment}
\usepackage[switch]{lineno}  %
\usepackage{graphicx}
\usepackage{multirow}
\usepackage[HTML]{xcolor}
\usepackage[export]{adjustbox}

\newcommand{\mb}{\mathbf}

\definecolor{green}{HTML}{00FF00}
\definecolor{red}{HTML}{FF0000}

\title{Subspace Modeling for Fast Out-Of-Distribution and Anomaly Detection}
\name{Ibrahima J. Ndiour, Nilesh A. Ahuja, Omesh Tickoo}
\address{Intel Labs}
\begin{document}
%\ninept
%
% \IEEEoverridecommandlockouts
% \IEEEpubid{\makebox[\columnwidth]{978-1-5386-5541-2/18/\$31.00~\copyright2018 IEEE \hfill} \hspace{\columnsep}\makebox[\columnwidth]{ }}
\maketitle
% \IEEEpubidadjcol
%
\begin{abstract}
This paper presents a fast, principled approach for detecting anomalous and out-of-distribution (OOD) samples in deep neural networks (DNN). We propose the application of linear statistical dimensionality reduction techniques on the semantic features produced by a DNN, in order to capture the low-dimensional subspace truly spanned by said features. We show that the \emph{feature reconstruction error} (FRE), which is the $\ell_2$-norm of the difference between the original feature in the high-dimensional space and the pre-image of its low-dimensional reduced embedding, is highly effective for OOD and anomaly detection. To generalize to intermediate features produced at any given layer, we extend the methodology by applying nonlinear kernel-based methods. Experiments using standard image datasets and DNN architectures demonstrate that our method meets or exceeds best-in-class quality performance, but at a fraction of the computational and memory cost required by the state of the art. It can be trained and run very efficiently, even on a traditional CPU. 

\end{abstract}
\begin{keywords}
Anomaly detection, out-of-distribution detection, uncertainty estimation, subspace modeling.
\end{keywords}
\section{Introduction}
\label{sec:intro}

Deep networks deployed in real-world conditions will inevitably encounter out-of-distribution data, which leads to outputs that are unpredictable, unexplainable and sometimes catastrophic. The ability to detect OOD data is therefore critical for the deployment of safe and transparent systems. 

OOD detection is typically performed by making the network provide an uncertainty score (along with the output) for each input. Early methods include the softmax score \cite{hendrycks2016baseline} and its temperature-scaled variants \cite{liang2018enhancing}. Bayesian neural networks \cite{gal2016dropout} and ensembles of discriminative classifiers \cite{lakshminarayanan2017simple} are more recent and can generate high quality uncertainty, but at the cost of complex model representations, and substantial compute and memory. Deep generative models learn distributions over the input data, and then evaluate the likelihood of new inputs with respect to the learnt distributions \cite{van2016conditional, hendrycks2019oe, ren2019likelihood}. Gradient-based characterization of abnormality in autoencoders is highlighted in \cite{9190706}. Finally, there are methods \cite{lee2018simple, ahuja2019bdl_dfm} that learn parametric class-conditional probability distributions over the features and use the likelihoods (w.r.t the learnt distributions) as uncertainty scores.
% Finally, there are methods \cite{lee2018simple, ahuja2019bdl_dfm} that involve learning parametric class-conditional probability distributions over the deep features in order to use the log-likelihoods (w.r.t the learnt distributions) as uncertainty scores.

Within the more general problem of OOD detection, anomaly detection has become particularly important in industrial applications. Its goal is to identify rare and abnormal events from the observation of data. Anomaly detection algorithms rely on good, defect-free samples during the training stage, and identify anomalous samples by comparing against the learned distribution of good data. Since this is really a specific type of OOD detection, state-of-the-art methods in anomaly detection are based on the same principles as the general OOD problem. For instance, \cite{differnet, ganomaly} use deep generative models to model the distribution of good samples. Alternately, methods such as \cite{defard2021padim, cohen2020sub} model clusters or distributions on a multi-level pyramid of deep-features. \cite{roth2021towards} uses greedy coreset subsampling (which is NP-hard) on the feature-banks to reduce memory requirements. These methods show impressive results to varying degrees, but at the cost of significant computational and storage complexity during training or inference. 

We present here a method for anomaly and OOD detection based on appropriately modeling the subspace of the intermediate features produced by a DNN. The high dimensionality of this feature space makes it very challenging, both computationally and algebraically, to perform a variety of otherwise routine tasks on the features, a phenomenon known as the  \emph{curse of dimensionality}. For instance, it leads to rank-deficiency in the data-matrix of the features.
 The \emph{manifold hypothesis} states, however, that real-world high-dimensional data lie on low-dimensional manifolds embedded within the high-dimensional space. \cite{tenenbaum2000global} prescribes that these high-dimensional spaces should be modeled by appropriate low-dimensional manifolds and sub-spaces. 
 In the computer vision field, the problems highlighted by the \emph{manifold hypothesis} are well understood for the image space: despite its very high dimensionality, many points in that space do not correspond to realistic natural images. In the context of the intermediate features of a DNN, this implies that the features sparsely occupy the high-dimensional space they live in. Hence, the true subspace spanned by the features can be accurately captured by appropriately mapping the original high-dimensional feature space to a reduced lower-dimensional subspace.
%  \cite{ahuja2019bdl_dfm} used principal component analysis (PCA) to model the lower-dimensional subspace of the deep features, following which density estimation was performed. 

 In this work, using a single-layer of a pretrained DNN, we show that the \emph{feature reconstruction error} (FRE), the $\ell_2$-norm of the difference between the original feature in the high-dimensional space and the pre-image of its low-dimensional reduced embedding, is a highly effective uncertainty score for anomaly and OOD detection. This circumvents the need to perform any subsequent processing in the reduced feature space, thereby greatly simplifying the procedure both during modeling and inference. We show that this approach achieves state-of-the-art anomaly and OOD detection performance (section \ref{subsec:OODresults}), but with significantly lower complexity compared to other methods. This makes it very attractive for deployment in real-world industrial usages on low-cost edge platforms without requiring investment into expensive discrete GPUs. Furthermore, the approach does not modify the network's parameters, which is a significant advantage for trained networks already deployed.

\begin{figure}
	\centering
	\includegraphics[width=0.49\textwidth]{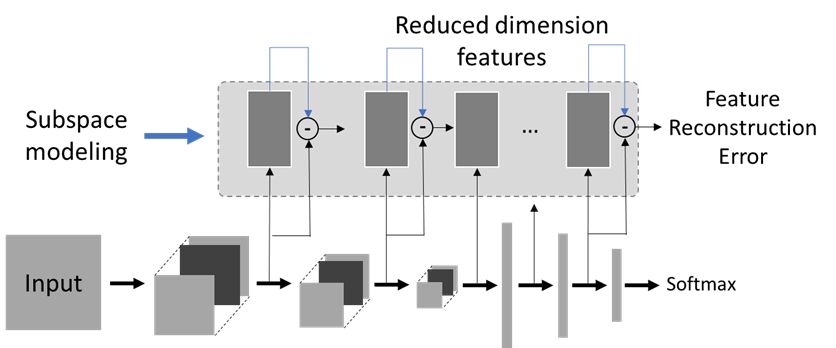}
	\caption{Proposed system.}
	\label{fig:system}
\end{figure}

\section{Approach}
\label{sec:approach}
Consider a deep neural network (DNN) trained on an $N$-class classification problem.
% Suppose we have a deep network trained to recognize samples from $N$ classes, $\{C_k\}, k=1, \ldots, N$. 
For an input $\mb{x}$, let $f(\mb{x})$ denote the output at an intermediate layer of the network. The features induced by the training dataset do not fully span the high-dimensional space in which they reside. Hence, for a training dataset of size $M$, the data-matrix $\mb{D} = [f(\mb{x}_1)|\dots|f(\mb{x}_M)] $ constructed from the features is rank deficient. Table \ref{table:data_ranks} shows how severe  the rank deficiencies in the higher-dimensional inner layers of a Resnet18 deep network used in our experiments are.
% , and how the sparsely-populated high dimensional feature space was meaningfully reduced.  
Hence, we learn a transformation $\mathcal{T}$ that maps the high-dimensional features onto an appropriate subspace, $\mathcal{T}: \mathcal{H} \to \mathcal{L}$ with $dim(\mathcal{L}) \ll dim(\mathcal{H})$. 
The parameters of the inverse transformation $\mathcal{T}^{\dagger}$ are also learnt simultaneously. 
% The subscript $i$ indicates that the transformation is learnt on a per layer basis. 
During inference, this transformation is applied to a test feature sample to obtain its reduced-dimension embedding.
% is embedded into the modeled subspace via the learnt subspace transformation. 
This reduced embedding is inverse-transformed into the original space and a \textit{feature reconstruction error} (FRE) score is calculated as the $\ell_2$ norm of the difference between the original and reconstructed vectors, as given by
% \begin{equation*}
%     FRE_i(\mb{x}) = \|f_i(\mb{x})-(\mathcal{T}_i^{\dagger} \circ \mathcal{T}_i)(f_i(\mb{x}))\|_2
% \end{equation*}
\begin{equation}
\label{eq:FRE}
    FRE(\mb{x}) = \|f(\mb{x})-(\mathcal{T}^{\dagger} \circ \mathcal{T})(f(\mb{x}))\|_2
\end{equation}
This score can be used as an uncertainty score for OOD detection. In what follows, we explain the various aspects of the subspace modeling process.
% using both linear and non-linear transformations. 
A complete flow-diagram of our approach is shown in Figure \ref{fig:system}.   

\subsection{Global vs Per-Class Subspace Modeling}
\begin{figure}
	\centering
	\includegraphics[width=0.43\textwidth]{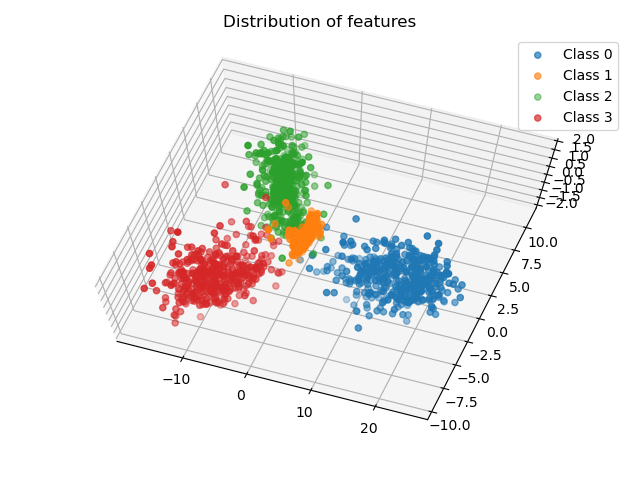}
	\caption{Distribution of features in 3D space. The features for Class 1 have spread mainly in the Z-dimension.}
	\label{fig:dist_3d}
\end{figure}

Subspace modeling can be applied either on the features from all classes of the training set at once, or on a per-class basis. In what follows, we refer to the former as global and the latter as per-class. Global modeling is appealing when the number of samples per class is small relative to the feature dimension, such as when dealing with a large number of classes but with limited training data per class. It is also an option when class labels are not available (semi-supervised OOD).
However, it may not adequately model the feature space's underlying structure as it does not take advantage of the fact that there may be multiple well-separated clusters, corresponding to separate classes or groups of classes. Modeling subspaces separately for each class can then often result in much better performance. This situation is clearly demonstrated in Figure \ref{fig:dist_3d} which shows the distribution of features from the penultimate layer of a simple CNN model trained with four classes. 

\begin{table}
    \footnotesize
	\caption{Feature dimensions and data-matrix ranks for Resnet18 trained on CIFAR10}
	\label{table:data_ranks}
	\centering
	\begin{tabular}{cccc}
		\toprule
		\textbf{Layer} & Layer 0 & Layer 1 & Layer 2\\
		\cmidrule(r){1-1}\cmidrule(r){2-4}
		Dimension & 512 & 256 & 512 \\
		Rank & 85 & 255 & 478 \\
		With $99.5\%$ PCA & 29 & 239 & 463 \\
		\bottomrule
	\end{tabular}
\end{table}

\subsection{Linear vs Non-Linear Subspace Modeling}
\textbf{Linear Subspace Modeling:} One popular choice for dimensionality reduction is principal component analysis (PCA) \cite{shlens2014tutorial}. In this framework, $\mathcal{H}$ and  $\mathcal{L}$ are, respectively, Euclidean spaces $\mathbb{R}^d$ and $\mathbb{R}^m$, with $m \ll d$. $\mathcal{T}$ 
% is then a linear transformation $\mathcal{T}: \mathbb{R}^d \to \mathbb{R}^m$, and 
can then be calculated from the singular value decomposition (SVD) \cite{golub2013matrix} of the data matrix $\mb{D}$. 
% The matrix representation of $\mathcal{T}$ is the truncation to the first $m$ columns of an orthogonal matrix obtained by juxtaposing the eigenvectors of the covariance matrix (ranked by descending order of eigenvalue). 
Table \ref{table:data_ranks} provides an intuition for the extent of dimensionality reduction achieved when applying PCA. For instance, we see that for Layer 0, the subspace dimension after applying PCA with $99.5$\% variability retention drops from $512$ to $29$, indicating that $99.5$\% of the information in the $512$-dimensional features is actually contained within a $29$-dimensional subspace! The results for other layers are less dramatic but nonetheless point to the need for appropriate subspace modeling. As the mapping from the high-dimensional feature space to the lower-dimensional reduced subspace is non-injective, there isn't a uniquely defined inverse image (pre-image) for a reduced feature i.e. $\mathcal{T}^{\dagger}$ is not uniquely defined. In the linear case, a common practice is to use the Moore-Penrose pseudo-inverse of the forward transformation \cite{golub2013matrix}.

\setlength\tabcolsep{2pt} % default value: 6pt
\begin{table}
\scriptsize
    % \footnotesize
% 	\caption{AUROC results for OOD detection performance. FRE scores are very competitive with other methods, often yielding the best performance despite their apparent simplicity.}
\caption{AUROC results for OOD detection performance.}
	\label{table:auroc}
	\centering
% 	\begin{tabular}{ccccccccc}
	\begin{tabular}{ccccccccc}
	
		\toprule
		\midrule
% 		\multirow{2}{*}{\begin{tabular}{c}
% 				CIFAR100  \\ (Wide-Resnet)
% 		\end{tabular}} 
% 		\multicolumn{2}{c}{CIFAR100} & \multicolumn{5}{c}{SVHN} & \multicolumn{5}{c}{LSUN}  \\
		 & \textbf{Mahal} & \textbf{LL} & \textbf{FRE} & \textbf{kFRE} &  \textbf{Mahal} & \textbf{LL} & \textbf{FRE} & \textbf{kFRE} \\
		\cmidrule(r){2-2}\cmidrule(r){3-3}\cmidrule(r){4-4}\cmidrule(r){5-5}\cmidrule(r){6-6}\cmidrule(r){7-7}\cmidrule(r){8-8}\cmidrule(r){9-9}		\textbf{CIFAR100} & \multicolumn{4}{c}{\textbf{SVHN (OOD)}} & \multicolumn{4}{c}{\textbf{LSUN (OOD)}}  \\
		\cmidrule(r){1-1}\cmidrule(r){2-5}\cmidrule(r){6-9}
		Layer2 & 91.5 & 92.8 & \textbf{93.1} & \textbf{93.4} & \textbf{98.5} & \textbf{98.8} & \textbf{98.3} & 98.2\\
		\cmidrule(r){1-1}\cmidrule(r){2-5}\cmidrule(r){6-9}
		Layer1 & 91.2 & 90.0 & 93.2 & \textbf{95.8} & \textbf{98.7} & \textbf{99.1} & 98.4 & 98.3\\
		\cmidrule(r){1-1}\cmidrule(r){2-5}\cmidrule(r){6-9}
		Layer0 & 75.0 & \textbf{84.8} & 79.3 & 75.9 & \textbf{97.3} & 95.1 & \textbf{97.1} & 91.5\\

		\cmidrule(r){1-1}\cmidrule(r){2-5}\cmidrule(r){6-9}
		Softmax & \multicolumn{4}{c}{74.3}  & \multicolumn{4}{c}{84.7}\\
		\midrule \midrule
		\textbf{CIFAR10} & \multicolumn{4}{c}{\textbf{SVHN (OOD)}} & \multicolumn{4}{c}{\textbf{LSUN (OOD)}}  \\
		\cmidrule(r){1-1}\cmidrule(r){2-5}\cmidrule(r){6-9}
		Layer2 & 94.6 & 94.5 & 77.2 & \textbf{98.5} & 98.8 & \textbf{99.4} & 95.3 & \textbf{99.0}\\
		\cmidrule(r){1-1}\cmidrule(r){2-5}\cmidrule(r){6-9}
		Layer1 & 86.4 & 88.8 & 48.5 & \textbf{92.4} & 72.5 & 86.0 & 65.2 & \textbf{87.0}\\
		\cmidrule(r){1-1}\cmidrule(r){2-5}\cmidrule(r){6-9}
		Layer0 & 95.2 & 95.0 & \textbf{96.7} & 93.9 & \textbf{95.1} & \textbf{95.5} & \textbf{95.3} & \textbf{95.1}\\
		\cmidrule(r){1-1}\cmidrule(r){2-5}\cmidrule(r){6-9}
		Softmax & \multicolumn{4}{c}{93.4}  & \multicolumn{4}{c}{94.0}\\
		\midrule \midrule
		\textbf{SVHN} & \multicolumn{4}{c}{\textbf{CIFAR10 (OOD)}} & \multicolumn{4}{c}{\textbf{LSUN (OOD)}}  \\
		\cmidrule(r){1-1}\cmidrule(r){2-5}\cmidrule(r){6-9}
		Layer2 & \textbf{94.2} & \textbf{93.8} & 85.2 & \textbf{93.7} & \textbf{94.3} & \textbf{93.9} & 90.1 & 93.5\\
		\cmidrule(r){1-1}\cmidrule(r){2-5}\cmidrule(r){6-9}
		Layer1 & 90.4 & \textbf{94.9} & 94.2 & \textbf{94.7} & 90.6 & \textbf{95.2} & 94.5 & \textbf{94.9}\\
		\cmidrule(r){1-1}\cmidrule(r){2-5}\cmidrule(r){6-9}
		Layer0 & 92.3 & \textbf{96.8} & 96.0 & 95.6 & 92.5 & \textbf{97.1} & 96.0 & 95.8 \\
		\cmidrule(r){1-1}\cmidrule(r){2-5}\cmidrule(r){6-9}
		Softmax & \multicolumn{4}{c}{93.0}  & \multicolumn{4}{c}{92.5}\\
		\midrule
		\bottomrule
	\end{tabular}
\end{table}

\noindent\textbf{Non-Linear Subspace Modeling:} Linear methods like PCA are most effective at subspace modeling if the underlying data is Gaussian, since PCA can only remove second-order dependencies \cite{shlens2014tutorial}. However, the assumption of normality for the intermediate features of a DNN was justified only at the penultimate layer \cite{lee2018simple}. For other layers, the distribution of features could significantly depart from the normal distribution. In general, it will depend on the dataset used to induce the feature set, as well as the network topology. In such situations, modeling the data as living in a lower-dimensional sub-manifold can yield vastly improved outcomes. Here, we use kernel PCA (kPCA) \cite{ham2004kernel} to model the underlying non-linear structure of the data. The choice of kPCA is motivated by the fact that it is a nonlinear extension of PCA, which allows for a direct comparison of performance between the linear and nonlinear OOD schemes. It is also computationally cheaper than other manifold learning methods such as Isomap \cite{tenenbaum2000global} and locally linear embedding (LLE) \cite{roweis2000nonlinear}, which can be described as kPCA on specially constructed Gram matrices \cite{ham2004kernel}. In general, though, our approach can employ any nonlinear manifold learning technique that provides an explicit mapping function for new data points. For computation of the inverse transformation, we refer to \cite{806} for a seminal paper on the topic. 

\section{Experiments and Results}
\label{sec:Results}

\subsection{Out-of-Distribution Detection}
\label{subsec:OODresults}
% \subsubsection{Experimental setup and evaluation metrics} 
\noindent\textbf{Experimental setup and evaluation metrics:}
% We first apply our method to the problem of OOD detection in deep networks. We use CIFAR10, CIFAR100 \cite{krizhevsky2009learning}, and SVHN \cite{netzer2011reading} as the in-distribution datasets. To test across networks of various depths and complexities, we train SVHN on Resnet20 ($0.27$M parameters), CIFAR10 on Resnet18 ($11.2$M parameters) \cite{He2016DeepRL}, and CIFAR100 on Wide-Resnet ($36.5$M parameters) \cite{zagoruyko2016wide}. For CIFAR10 and CIFAR100, we use SVHN and a resized version of LSUN \cite{yu15lsun} as OOD datasets. For SVHN as the in-distribution dataset, CIFAR10 and LSUN are used as OOD datasets. In all experiments, the subspace transformations are estimated from the training split of the in-distribution dataset, while performance metrics are calculated on the test splits. We apply both the linear (PCA) and nonlinear (kPCA with RBF kernel) subspace techniques to model the feature subspace on a per-class basis, with results reported separately in Table \ref{table:auroc}. Additionally, we perform our experiments on three layers of the networks listed above, with layers chosen to be located uniformly along the network path. The layers are labelled as 0, 1, and 2, with 0 being the (semantic) outermost layer, and 1, 2 being progressively deeper within the network. 

For the problem of OOD detection, we use CIFAR10, CIFAR100, and SVHN as the in-distribution datasets. To test across networks of various depths and complexities, we train SVHN on Resnet20 ($0.27$M parameters), CIFAR10 on Resnet18 ($11.2$M parameters), and CIFAR100 on Wide-Resnet ($36.5$M parameters). For models trained on CIFAR10 and CIFAR100, we use SVHN and a resized version of LSUN as OOD datasets; for those trained on SVHN, CIFAR10 and LSUN are used as OOD datasets. In all experiments, the subspace transformations are estimated from the training split of the in-distribution dataset, while performance metrics are calculated on the test splits. We apply both the linear (PCA) and nonlinear (kPCA with RBF kernel) subspace techniques to model the feature subspace on a per-class basis, with results reported separately in Table \ref{table:auroc}. We tested our method on three layers of each of the networks, with layers chosen to be located uniformly along the network path. The layers are labelled as 0, 1, and 2, with 0 being the (semantic) outermost layer, and 1, 2 being progressively deeper within the network. 

\begin{figure}
	\centering
	\includegraphics[width=0.23\textwidth]{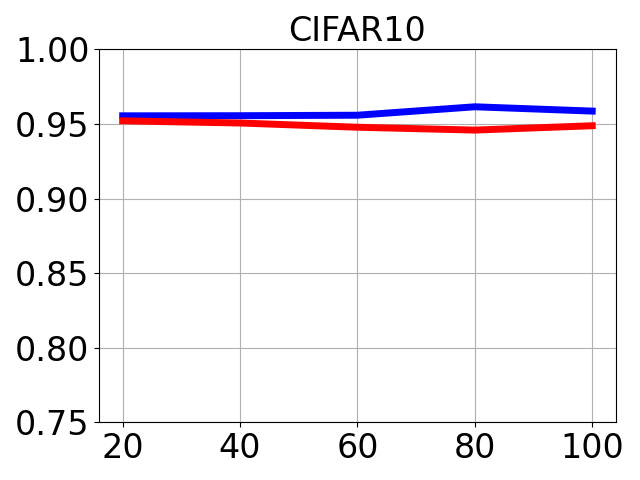}
	\includegraphics[width=0.23\textwidth]{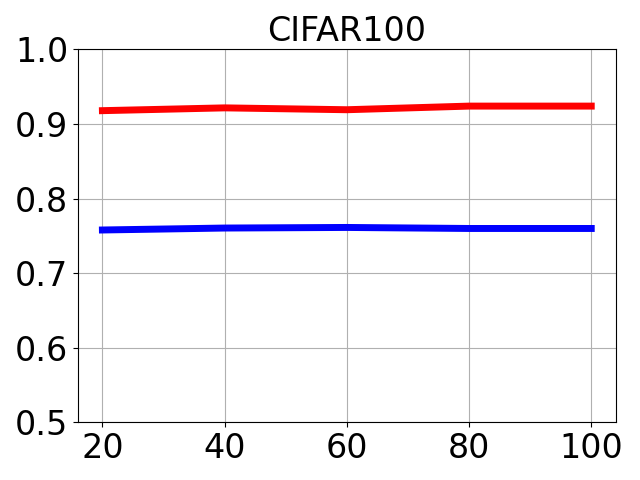}
	\caption{AUROC (Y-axis) for OOD detection with CIFAR10 and CIFAR100 as in-distribution datasets, and SVHN (red) and LSUN (blue) as OOD sets as we decrease the percentage of training data used (X-axis) for our OOD method all the down to 20\%. There is virtually no degradation in performance. }
	\label{fig:robustness}
\end{figure}

\begin{figure}
	\centering
	\includegraphics[width=0.10\textwidth, cfbox=green 1.7pt 0pt]{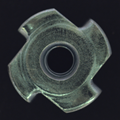}
	\includegraphics[width=0.10\textwidth, cfbox=red 1.7pt 0pt]{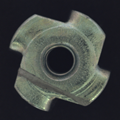}
	\quad
	\includegraphics[width=0.10\textwidth, cfbox=green 1.7pt 0pt]{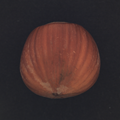}
	\includegraphics[width=0.10\textwidth, cfbox=red 1.7pt 0pt]{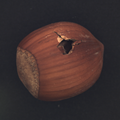}
	\caption{Good (green) and defective (red) samples from the MVTec dataset.}
	\label{fig:mvtec}
\end{figure}

During testing, the FRE (Eq. (\ref{eq:FRE})) is used to distinguish between in distribution and out-of-distribution data. This effectively creates a binary classifier, whose performance is characterized by the receiver operating characteristics (ROC) curve and we report the area under the ROC curve (AUROC) in Table \ref{table:auroc}. For baseline comparison, we use a binary OOD classifier based on Softmax scores (applicable only at the Softmax layer). At each layer tested, we also benchmark against a binary classifier based on Mahalanobis scores (denoted as Mahal). The Mahalanobis scores are the likelihoods of test samples w.r.t feature probability distributions fitted with multivariate Gaussian distributions sharing the same covariance across classes. In Table \ref{table:auroc}, FRE (resp. kFRE) refers to the feature reconstruction error with PCA (resp. kPCA) and LL corresponds to results obtained with \cite{ahuja2019bdl_dfm}. 
% We deem any set of AUROC results that are within a half-percentage point of each other to be a statistical tie. 

% Finally, note that although we have investigated the OOD detection performance on a per-layer basis, it is possible to combine all such per-layer scores in order to improve the overall detection performance. For instance, the scores could be combined using any binary classifier such as logistic regression trained using a hold-out auxiliary OOD data set.

% \subsubsection{Subspace modeling} In Table \ref{table:data_ranks}, we show the ranks of the data-matrices at various layers of Resnet18 trained on CIFAR10. For each layer, the rank is much lower than the feature dimension, confirming that the features actually reside in a lower-dimensional subspace. An intuition for the extent of dimensionality reduction possible can be obtained by applying PCA when retaining a very high amount of data variability. For instance, at Layer $0$, while the original feature dimension is $512$, the rank indicates that the actual subspace dimension can be at most $85$. Further, applying PCA with $99.5$\% variability retention reduces its dimension to $29$, i.e. $99.5$\% of the information in the $512$-dimensional features is actually contained within a $29$-dimensional subspace! The results for other layers similarly point to the need for appropriate subspace modeling. 
\begin{table}
% \scriptsize
    \scriptsize
	\caption{Anomaly Detection AUROC MVTec dataset \\{\footnotesize $^\star$ Methods not reporting class-itemized results. For PaDim, results were recreated using \emph{anomalib}\cite{anomalib}}}
% 	\caption{AUROC results for anomaly detection on MVTec (^\star for methods not reporting class-itemized results}
	\label{table:mvtec}
	\centering
	\begin{tabular}{ccccccc}
	
		\toprule
		\midrule
		 Category & GANomaly & DifferNet & SPADE$^\star$ & PaDim$^\star$ & PatchCore & FRE (Ours) \\
		\midrule
    Carpet & 69.9 & 92.9 & - & 99.5 & 98.7 & 100 \\
		\midrule
    Grid & 70.8 & 84.0 & - & 94.2 & 98.2 & 95.8 \\
		\midrule
    Leather & 84.2 & 97.1 & - & 100 & 100 & 100 \\
		\midrule
    Tile & 79.4 & 99.4 & - & 97.4 & 98.7 & 97.8 \\
		\midrule
    Wood & 83.4 & 99.8 & - & 99.3 & 99.2 & 99.4 \\
		\midrule
    Bottle & 89.2 & 99.0 & - & 99.9 & 100 & 100 \\
		\midrule
    Cable & 75.7 & 95.9 & - & 87.8 & 99.5 & 99.3 \\
		\midrule
    Capsule & 73.2 & 86.9 & - & 92.7 & 98.1 & 99.4 \\
		\midrule
    Hazelnut & 78.5 & 99.3 & - & 96.4 & 100 & 99.8 \\
		\midrule
    Metal Nut & 70.0 & 96.1 & - & 98.9 & 100 & 96.9 \\
		\midrule
    Pill & 74.3 & 88.8 & - & 93.9 & 96.6 & 95.7 \\
		\midrule
    Screw & 74.6 & 96.3 & - & 84.5 & 98.1 & 97.5 \\
		\midrule
    Toothbrush & 65.3 & 98.6 & - & 94.2 & 100 & 99.4 \\
		\midrule
    Transistor & 79.2 & 91.1 & - & 97.6 & 100 & 98.6 \\
		\midrule
    Zipper & 74.5 & 95.1 & - & 88.2 & 99.4 & 96.4 \\
		\midrule
    \textbf{Average} & \textbf{76.2} & \textbf{94.9} & \textbf{85.5} & \textbf{97.9} & \textbf{99.1} & \textbf{98.4} \\
		\bottomrule	
	\end{tabular}
\end{table}

% \subsubsection{OOD detection results} 
\noindent\textbf{OOD detection results:}
Table \ref{table:auroc} shows that the proposed method is very competitive, often outperforming benchmarks that are much more demanding in computations and memory storage, and typically within a half-percentage point of those. In particular, both benchmarks require density estimation and likelihood evaluation in high-dimensional spaces, while the proposed method relies on a few simple dot-product operations in the linear case. Of note, we notice a trend of the nonlinear scheme providing better results than its linear counterpart as we progress deeper into the network. This is consistent with the hypothesis alluded to earlier that 
% As alluded to earlier, the nature of the feature space is dependent on the network topology and the training images used to induce it. In our observations, 
the outer-most layers produce features with a distribution close to Gaussian while deeper layers have feature spaces that may exhibit complex nonlinearity in their structure.

\begin{table}
% \scriptsize
    \scriptsize
	\caption{Anomaly Detection AUROC on Magnetic Tile}
	\label{table:mtd}
	\centering
	\begin{tabular}{ccccc}
	
		\toprule
		\midrule
		 GANomaly & I-NN & DifferNet & PatchCore & FRE (Ours) \\
 		\\
		 
		\midrule
     76.6 & 80.0 & 97.7 & 97.9  & 99.2 \\
		\bottomrule
	\end{tabular}
\end{table}

% \subsubsection{Robustness to reduced training data} 
\noindent\textbf{Robustness to reduced training data:}
In practical situations, we might not have access to the entire training dataset. We show that our method remains very effective at OOD detection even when its training (subspace modeling) is performed with a fraction of the training data. The plots in Figure \ref{fig:robustness} show the variations in AUROC for the CIFAR100 dataset, using both linear and non-linear subspace modeling, as the percentage for training data is gradually reduced to $20$\%. We see that the performance remains very stable, showing a decrease of less than $1$\% in AUROC scores in the majority of cases. We observe this trend for the other datasets as well.

\subsection{Application to Anomaly Detection}
\label{subsec:ADresults}
Finally, we apply our method to the problem of anomaly detection in images. 
% This is useful, for instance, in industrial setups to automatically detect defective parts on a manufacturing line. 
We learn a linear PCA transform, $\mathcal{T}$, on the features of an EfficientNet-B5 pretrained on Imagenet from only the defect-free images. We then use the feature-reconstruction score to distinguish between good and defective samples. We test our approach on two datasets: the MVTec anomaly detection dataset \cite{bergmann2019mvtec} (sample images shown in Figure \ref{fig:mvtec}), and the Magnetic Tile defect (MTD) dataset \cite{huang2020surface}. The results, comparing against various available benchmarks, are shown in Tables \ref{table:mvtec} and \ref{table:mtd} respectively. Our method attains the best performance on MTD and a close second on MVTec, despite its simplicity.

\noindent\textbf{Complexity:} Existing state-of-the-art methods involve significant complexity during training or inference. \cite{differnet, ganomaly} use deep generative models (GANs or normalizing flows) to model the distribution of normal samples, which require expensive training. While \cite{cohen2020sub, defard2021padim} mititgate training complexity by using pretrained models, they both involve modeling clusters or probability distributions on a multi-level pyramid of deep-features. 
% In particular, \cite{defard2021padim} involves storing the parameters of learnt multivariate Gaussians for each image patch, which requires a large amount of storage. 
\cite{roth2021towards} reduces the memory storage requirements of feature-pyramid based methods but uses greedy coreset subsampling on the stored feature banks to accomplish this, which is known to be a computationally involved process (NP-hard). By contrast, our method does not involve training a new model of any kind (discriminative or generative), operates on features from a single layer of the deep-network (instead of a feature pyramid), and does not involve any complex probabilistic modeling. It achieves state-of-the-art performance with remarkably low computational overhead, making it very attractive to deploy in real-world industrial usages.

\section{Conclusion}
\label{sec:conclusion}
% Deep networks have become ubiquitous in image processing and computer vision. However, their safe deployment requires practical and reliable uncertainty measures for anomaly and out-of-distribution detection.
% % , which remain elusive as of now. 
% This work sketches initial progress with the application of linear statistical dimensionality reduction on the semantic features of a DNN, prior to leveraging the \emph{feature reconstruction error} as an uncertainty score. The method is simple, principled and very fast. Our experimentation shows qualitative performance at par or better than other state-of-the-art methods that are significantly more complex. Generalizing to intermediate features produced at any given layer, we further extended the method with nonlinear subspace modeling to account for the presence of nonlinearity in the feature space. 
% % In future work, we will seek to combine the individual uncertainty scores estimated at each layer.
This work sketched initial progress on anomaly and OOD detection with the application of linear and nonlinear dimensionality reduction on the semantic features of a DNN, prior to leveraging the \emph{feature reconstruction error} as an uncertainty score. The method is simple, principled and very fast. Experimentations show qualitative performance at par or better than state-of-the-art methods that are significantly more complex.

% References should be produced using the bibtex program from suitable
% BiBTeX files (here: strings, refs, manuals). The IEEEbib.bst bibliography
% style file from IEEE produces unsorted bibliography list.
% -------------------------------------------------------------------------
\bibliographystyle{IEEEbib}
\bibliography{strings,refs}

\end{document}